\documentclass[10pt,conference]{IEEEtran}
\usepackage[utf8]{inputenc} 
\usepackage[T1]{fontenc}
\usepackage{url}
\usepackage{ifthen}
\usepackage{cite}
\usepackage[cmex10]{amsmath} 
\usepackage{amsmath,amsfonts}

\usepackage{array}
\usepackage{textcomp}
\usepackage{stfloats}
\usepackage{url}
\usepackage{verbatim}
\usepackage{graphicx}
\usepackage{cite}
\usepackage{multicol}

\usepackage{algpseudocode}
\usepackage{bm}
\usepackage{verbatim}
\usepackage{booktabs}
\usepackage{amsmath,amssymb,amsfonts}
\usepackage{threeparttable}
\usepackage{multirow}
\usepackage{stackengine}

\usepackage[caption=false,font=normalsize,labelfont=sf,textfont=sf]{subfig}

\usepackage{hyperref}
\hypersetup{colorlinks,
	linkcolor=blue,%
	citecolor=green}

\usepackage{xspace} 

\makeatletter
\DeclareRobustCommand\onedot{\futurelet\@let@token\@onedot}
\def\@onedot{\ifx\@let@token.\else.\null\fi\xspace}

\def\etal{\emph{et al}\onedot}
\makeatother

\usepackage{enumitem}
\usepackage[linesnumbered,ruled,vlined]{algorithm2e}
\usepackage{graphicx}  

\IEEEoverridecommandlockouts

\begin{document}

\title{\huge MOC-RVQ: Multilevel Codebook-Assisted Digital Generative Semantic Communication}

\author{
Yingbin Zhou$^{1,2}$~
Yaping Sun$^{3,1}$~
Guanying Chen$^{1}$~
Xiaodong Xu$^{4,3}$\\
Hao Chen$^3$~
Binhong Huang$^3$~
Shuguang Cui$^{2,1,3}$~
Ping Zhang$^{4,3}$\\
$^1$FNii-Shenzhen, CUHKSZ, Shenzhen, China ~
$^2$SSE, CUHKSZ, Shenzhen, China \\
$^3$Pengcheng Laboratory, Shenzhen, China \\
$^4$Beijing University of Posts and Telecommunications, Beijing, China

\thanks{

This work was supported in part by NSFC with Grant No. 62293482 and Grant No. 62301471, the Basic Research Project No. HZQB-KCZYZ-2021067 of Hetao Shenzhen-HK S\&T Cooperation Zone, the Shenzhen Outstanding Talents Training Fund 202002, the Guangdong Research Projects No. 2017ZT07X152 and No. 2019CX01X104, the Guangdong Provincial Key Laboratory of Future Networks of Intelligence (Grant No. 2022B1212010001), and the Shenzhen Key Laboratory of Big Data and Artificial Intelligence (Grant No. ZDSYS201707251409055), and the Major Key Project of PCL Department of Broadband Communiation (PCL2023AS1-1).

The open source code is provided at: \href{https://github.com/Albert2X/moc_rvq}{https://github.com/Albert2X/moc\_rvq}
}
}

\maketitle

\begin{abstract}
Vector quantization-based image semantic communication systems have successfully boosted transmission efficiency, but face challenges with conflicting requirements between codebook design and digital constellation modulation. Traditional codebooks need wide index ranges, while modulation favors few discrete states. To address this, we propose a multilevel generative semantic communication system with a two-stage training framework. In the first stage, we train a high-quality codebook, using a multi-head octonary codebook (MOC) to compress the index range. In addition, a residual vector quantization (RVQ) mechanism is also integrated for effective multilevel communication. In the second stage, a noise reduction block (NRB) based on Swin Transformer is introduced, coupled with the multilevel codebook from the first stage, serving as a high-quality semantic knowledge base (SKB) for generative feature restoration. Finally, to simulate modern image transmission scenarios, we employ a diverse collection of high-resolution 2K images as the test set. The experimental results consistently demonstrate the superior performance of MOC-RVQ over conventional methods such as BPG or JPEG. Additionally, MOC-RVQ achieves comparable performance to an analog JSCC scheme, while needing only one-sixth of the channel bandwidth ratio (CBR) and being directly compatible with digital transmission systems.

\end{abstract}


\section{Introduction}
\label{sec:intro}

Propelled by the rapid development of deep learning technology, a new intelligent transmission paradigm known as semantic communication is gradually gaining prominence. In contrast to traditional transmission, semantic communication aims for the transmission of semantic fidelity~\cite{bao2011towards}. For example, Dai \etal~\cite{dai2022nonlinear} integrate the nonlinear transform as a robust prior to efficiently extract source semantic features for image transmission. Sun \etal~\cite{sun2023semantic} design a multi-level semantic coding and feature transmission mechanism powered by semantic knowledge base. Wu \etal~\cite{wu2023cddm} propose channel denoising diffusion models (CDDM) for wireless communications by developing specific training and sampling algorithms tailored to the forward diffusion process and the reverse sampling process. However, these works suffer from efficiency issues, due to the high dimension of latent representation of raw data.
 
One solution to achieve compact representation is vector quantization (VQ) technique~\cite{van2017neural} . Generally, VQ transforms semantic features into a series of indices, providing a more compact format that can be further converted into bits for transmission, then the receiver utilizes the received indices to reconstruct the semantic features through a pre-trained semantic codebook. Ultimately, the raw data or the task-oriented information can be decoded from these reconstructed semantic features. For instance, Nemati \etal~\cite{nemati2022all} delve into the characteristics of VQ-VAE and adapt its training process to formulate a robust JSCC scheme against noisy wireless channels. Fu \etal~\cite{fu2023vector} design a CNN-based transceiver to extract multi-scale semantic features and incorporate multi-scale semantic embedding spaces to facilitate feature quantization. 

However, the VQ methods mentioned above \textbf{suffer the following two issues}:

\begin{itemize}

\item \textbf{Incompatible issue} arises from the disparity between conventional vector quantization and digital constellation modulation. The former, aimed at achieving optimal image representation, normally employs a learnable codebook with wide index ranges. In contrast, the latter is more inclined to handle fewer states (e.g., 16-QAM, 64-QAM, etc.). While it is conceivable to regroup bits representing indices to align with modulation, such an adjustment also disrupts the explicit relationship between indices and constellation points. Note that this explicit correspondence is crucial for preserving the local semantic relationship of the underlying neighbour code vectors, which directly affects the quality of reconstructed image.

\item \textbf{Mismatch issue} occurs in the local relationship between code indices and code vectors, which renders vector quantization-based semantic communication systems susceptible to channel noise. For instance, while the difference between index "1" and index "2" is 1, the distance between the underlying code vectors of index "1" and index "2" can be substantial. This mismatch may lead to errors in decoding and, consequently, degrade the overall performance of the communication system. 

\end{itemize}

\textbf{To tackle the \textit{incompatible issue}}, we introduce MOC-RVQ with a two-stage training framework.
In Stage 1, we introduce: (1) MOC, a variant of vector quantizer by constructing a multi-head octonary codebook (MOC) to compress the range of indices to a limit of 8, \textbf{allowing for a direct match with 64-QAM.} (2) RVQ, a multi-level semantic transmission mechanism based on residual vector quantization (RVQ), which is helpful to compensate the negative effect of the quantization noise~\cite{lee2022autoregressive,yang2023hifi}. We combine MOC and RVQ to form MOC-RVQ. 
In Stage 2, we introduce a noise reduction block (NRB) along with feature requantization to enable generative feature restoration.
\textbf{For the \textit{mismatch issue}}, we adopt a heuristic codebook reordering algorithm, which is aimed at \textbf{minimizing the semantic distance} between underlying code vectors with nearby indices. 


Overall, \textbf{the main contributions are summarized as follows:}

\begin{itemize}[itemsep=0pt,parsep=0pt,topsep=2bp]
    \item We point out two inherent shortcomings of the existing VQ-based semantic communication systems and propose a novel two-stage training framework. 
    
    \item We design MOC-RVQ to align with the preference of digital constellation modulation. Then the noise reduction block (NRB) based on Swin Transformer and feature requantization are incorporated for generative semantic reconstruction.
    
    \item We develop a codebook reordering algorithm to mitigate inconsistencies in the local relationships among code vectors with adjacent indices. This algorithm serves to enhance the overall robustness of the system against channel noise.
        
\end{itemize}

\section{Preliminary}
\label{sec:preliminary}

\begin{figure}[t] \centering
    \includegraphics[width=0.48\textwidth,height=0.3\textwidth]{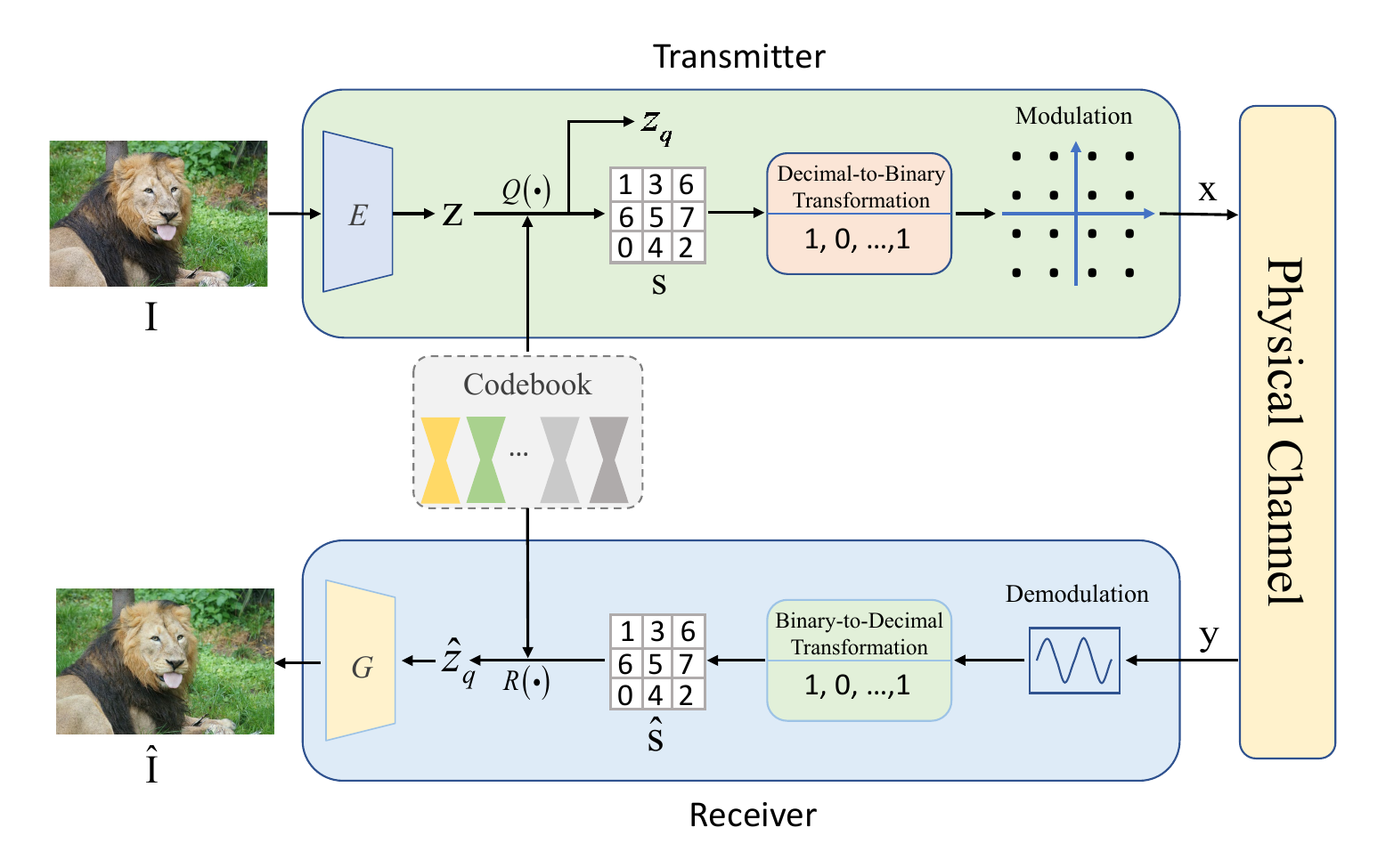}
    \caption{An essential structure for VQ-based semantic communication with the presence of a stochastic physical channel.} \label{fig:vq_pipeline}
\end{figure}

In this section, we present a general VQ-based semantic communication framework as preliminary. Note that this framework will be expanded and adjusted into our proposed two-stage framework in Section~\ref{sec:method}. The main process is illustrated in Fig.~\ref{fig:vq_pipeline}. In this design, both the transmitter and receiver utilize a shared pre-trained semantic codebook to efficiently compress and reconstruct semantic features, thereby reducing communication overhead.

\subsection{Transmitter}%
\label{sub:Transmitter}

Given an input image $\mathbf{I} \in \mathbb{R}^{H \times W \times 3}$, the semantic encoder $E$ (serves as understanding knowledge base) is employed initially to transform $\mathbf{I}$ into semantic feature $\mathbf{z} \in \mathbb{R}^{h \times w \times n_q}$. Subsequently, each grid feature $\mathbf{z}^{(i,j)} \in \mathbb{R}^{n_q}$ is substituted with the nearest vector in the learnable codebook $\mathcal{C}=\left\{\mathbf{e}_{k} \in \mathbb{R}^{n_q}\right\}_{k=0}^{N}$. This substitution yields the quantized feature $\mathbf{z}_{q}$ along with the corresponding code index sequence $\mathbf{s} \in \{0,1, \cdots, N-1\}^{h \cdot w}$, by the following equations:
\begin{align}
\mathbf{s}^{(i,j)} & = \underset{k}{\operatorname{argmin}}\left\|\mathbf{z}^{(i,j)}-\mathbf{e}_{k}\right\| \in \{0,1, \cdots, N-1\},\\
\mathbf{z}_{q}^{(i,j)} & = \mathbf{Q}\left(\mathbf{z}^{(i,j)}\right) = \mathbf{e}_{\mathbf{s}^{(i,j)}} \in \mathbb{R}^{n_q},
\end{align}

where, $\mathbf{Q}(\cdot)$ represents a quantization operation that maps the continuous encoded grid feature $\mathbf{z}^{(i,j)}$ to its quantized form $\mathbf{z}_{q}^{(i,j)}$ by selecting the closest codebook vector $\mathbf{e}_k$. Notably, \textbf{the code index sequence $\mathbf{s}$ can serve as a compact equivalent representation of the quantized feature $\mathbf{z}_{q}$}, meaning that transmitting only the code index sequence $\mathbf{s}$ can effectively reduce communication overhead. After applying decimal-to-binary transformation and digital constellation modulation, $\mathbf{s}$ is transformed into a transmitted complex signal $\mathbf{x} \in \mathbb{C}^{B \times 1}$, where $B$ denotes the length of the transmitted signal.

\begin{figure*}[t] \centering

    \includegraphics[width=0.90\textwidth,height=0.27\textwidth]{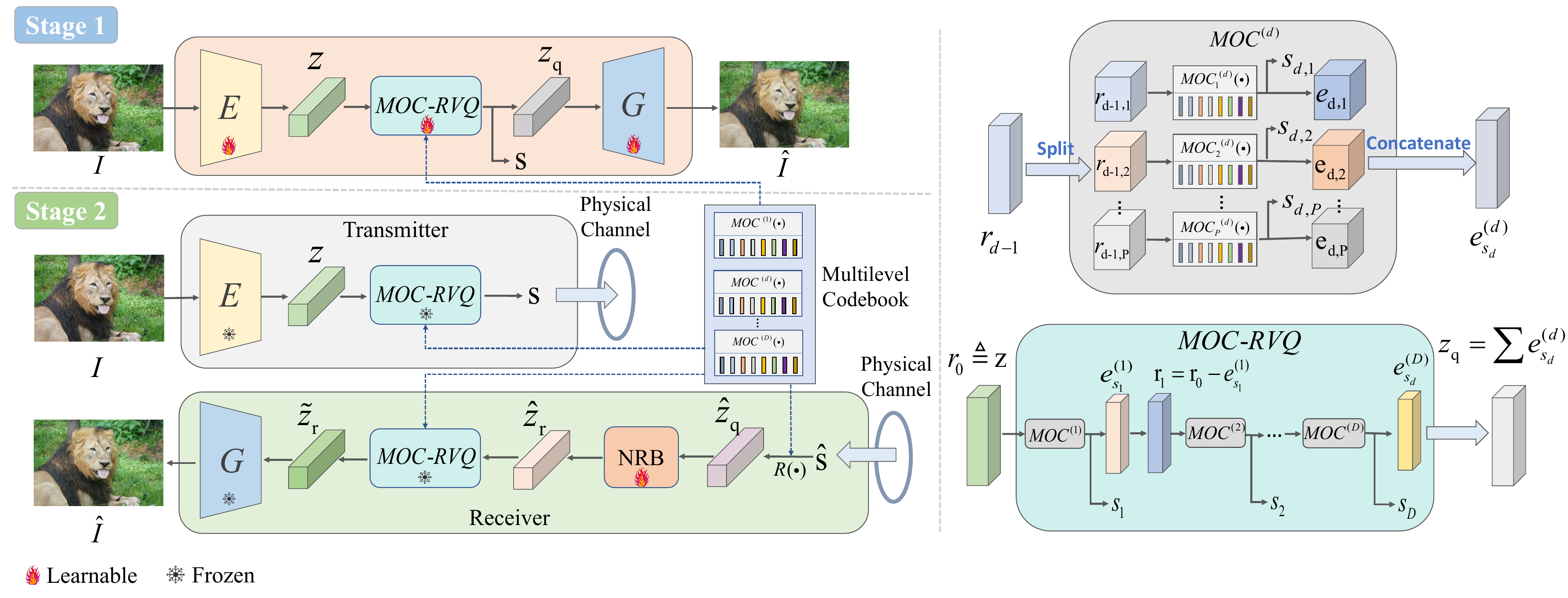}
    \caption{\textbf{Left:} The proposed two-stage framework. In Stage 1, we initially pretrain MOC-RVQ and other model components to generate compact representation of high-resolution image. Then the noise reduction block (NRB) is finetuned in Stage 2 to achieve feature restoration. \textbf{Right:} The architecture of the proposed MOC and MOC-RVQ. In MOC, each residual feature $\mathbf{r}_{d-1}$ is first divided into several heads, then the 8-state codebooks are employed to quantize each head feature, and finally these features are concatenated together to form a single quantized feature. In MOC-RVQ, feature are recursive quantized by MOCs to produce residual features and the corresponding code indices. Note that the output of MOC-RVQ is a summation over all quantized features $\mathbf{e}^{(d)}_{\mathbf{s}_d} $.}
    \label{fig:framework}
\end{figure*}

\subsection{Physical Channel}%
\label{sub:Physical_Channe}
To simplify channel modeling without sacrificing representativeness, we adopt the additive white Gaussian noise (AWGN)  as follows:
\begin{equation}
\mathbf{y} = \mathbf{x} + \mathbf{w},
\end{equation}
where $\mathbf{y} \in \mathbb{C}^{B \times 1}$ represents the received complex signal. The AWGN $\mathbf{w}$ has i.i.d elements with zero mean and variance $\sigma^2$.

\subsection{Receiver}%
\label{sub:Receiver}

Following the demodulation of the received signal $\mathbf{y}$ and the subsequent binary-to-decimal transformation, the receiver retrieves the indices $\mathbf{\hat{s}}$ from $\mathbf{y}$. It is noteworthy that $\mathbf{\hat{s}}$ can be analogized to the prompt in large language models, and we employ it to generate high-quality reconstructed images. Specifically, by \textbf{selecting the corresponding embedding vectors from the codebook $\mathcal{C}$ using the received indices $\mathbf{\hat{s}}$,} the receiver reconstructs the semantic feature $\mathbf{\hat{z}}_q$, we define this process as semantic feature retrieval $\mathbf{R}(\cdot)$. Then, leveraging the retrieval semantic feature $\mathbf{\hat{z}}_q$, the semantic decoder $G$ (serves as generative knowledge base) generates the reconstructed image $\mathbf{\hat{I}}$.

\section{The Proposed Method}%
\label{sec:method}

In this section, we expand the pipeline mentioned in Section~\ref{sec:preliminary} and introduce a novel two-stage training framework to enhance VQ-based semantic communication. Subsequently, we delve into the description of codebook reordering, an algorithm employed to minimize the mismatch between local relationships of code indices and code vectors.

\subsection{Stage 1: Pretraining of MOC-RVQ}%
\label{sub:MOC}

The primary objective of Stage 1 is to train a high-quality encoder/decoder model and multilevel codebook \textbf{without any channel noise}, aiming for a compact representation of image. As depicted in Fig.~\ref{fig:framework} (Left), in Stage 1, an input image $\mathbf{I} \in \mathbb{R}^{H \times W \times 3}$ is initially fed into the encoder $E$ to yield a continuous semantic feature $\mathbf{z} \in \mathbb{R}^{h \times w \times n_q}$. Subsequently, the MOC-RVQ layer quantizes this feature, resulting in the quantized feature $\mathbf{z}_{q}$ and the associated code indices sequence $\mathbf{s} \in \{0, \cdots, N-1\}^{h \cdot w \cdot P \cdot D}$. Here, $P$ and $D$ represent the number of heads and the level of quantizations respectively, and \textbf{$N$ is set to 8 to directly align with the few states preference of 64-QAM.} Finally, the quantized feature $\mathbf{z}_{q}$ is fed into the decoder $G$ to reconstruct the image as $\mathbf{\hat{I}}$.

The detailed designs of multi-head octonary codebook (MOC) and MOC-RVQ are illustrated in Fig.~\ref{fig:framework} (Right), and their functionality can be described as follows:

\paragraph{MOC} Given a quantization level $d$, each residual feature $\mathbf{r}_{d-1}\in \mathbb{R}^{h \times w \times n_q}$ is initially divided into several heads $\mathbf{r}_{d-1,i} \in \mathbb{R}^{h \times w \times n_q/P}$ (typically set $P$ to 4 by default). Subsequently, the 8-state codebooks are employed to quantize each head separately, resulting in head-wise quantized features $\mathbf{e}_{d,i}$. These quantized features are then concatenated to form a single quantized feature $\mathbf{e}_{s_d}^{(d)}$. We name this distinctive codebook design for aliging the preference of digital constellation modulation as \textbf{multi-head octonary codebook (MOC).} It is worth noting that, unlike a conventional codebook of size $K$, our proposed MOC significantly expands the feature matching space by generating $N^P$ discrete states through the combination of individual head states.

\paragraph{MOC-RVQ} By replacing the naive quantizer with MOC, we introduce a multilevel semantic transmission mechanism based on residual vector quantization (RVQ) to counteract potential adverse effects resulting from the quantization process, ultimately further enhances the system's reconstruction performance. Specifically, with a default quantization level $D$ set to 4, the semantic feature $z$ undergoes recursive quantization to produce residual code indices as follows:

\begin{equation}
\mathcal{MOC-RVQ}(\mathbf{z}) = \left(\mathbf{s}_{1}, \cdots, \mathbf{s}_{d}, \cdots, \mathbf{s}_{D}\right),
\end{equation}
where $\mathbf{s}_d \in \{0, \cdots, N-1\}^{h \cdot w \cdot P}$ represents the code indices of $\mathbf{z}$ at level $d$, and $P$ denotes the number of heads. More specifically, given an initial residual $\mathbf{r}_0=\mathbf{z}$, \textbf{the recursive process computes indices $s_d$ and the next residual $\mathbf{r}_d$ as follows:}

\begin{align}
\mathbf{s}_d & = \mathcal{MOC}^{(d)}(\mathbf{r}_{d-1}), \\
\mathbf{r}_d & = \mathbf{r}_{d-1} - \mathbf{e}^{(d)}_{\mathbf{s}_d},
\end{align}
where $\mathcal{MOC}^{(d)}(\cdot)$ is the $d$-th MOC quantizer, and $\mathbf{e}^{(d)}$ is the corresponding codebook embedding. The final quantized result of MOC-RVQ is obtained by $\mathbf{z}_q = \sum_{d=1}^{D} \mathbf{e}^{(d)}_{\mathbf{s}_d}$.

\paragraph{Training Objectives}
To train the proposed model, several loss functions are introduced into our method. Since the feature quantization operation is non-differentiable, we adopt the approach from ~\cite{esser2021taming} by copying gradients from $G$ to $E$ during backpropagation. This enables end-to-end training of the encoder/decoder model and codebook, guided by the following objective function:
\begin{align}
\mathcal{L}^{\prime}_{V Q}(E, G, \mathcal{C})=\left\| \mathbf{\hat{I}}-\mathbf{I}\right\|_{1}+\sum_{d,h}\|\operatorname{sg}[\mathbf{r}_{d-1,h}]-\mathbf{e}_{d,h}\|_{2}^{2},
\end{align}
where $sg[\cdot]$ represents the stop-gradient operation. Note that the first term of $\mathcal{L}^{\prime}_{VQ}$ stands for the image-level $L_1$ loss, while the second term is employed for updating the proposed multilevel octonary codebooks.

To enhance texture restoration, a semantic guidance loss, as proposed in ~\cite{chen2022real}, is also introduced as a regularization term:
\begin{align}
\mathcal{L}_{V Q}=\mathcal{L}_{V Q}^{\prime}+\gamma\|\operatorname{CONV}(\mathbf{z})-\phi(\mathbf{I})\|_{2}^{2},
\end{align}
where $CONV(\cdot)$ denotes a $1 \times 1$ convolution layer to align the dimensions of $\mathbf{z}$, $\phi$ represents the pretrained VGG19 model, and $\gamma$ is set to 0.1. Additionally, in accordance with~\cite{esser2021taming}, perceptual loss and adversarial loss are also employed for model pretraining.

\subsection{Stage 2: Finetuning for Noise Reduction}%
\label{sub:Finetuning}
Since Stage 1 is trained end-to-end without considering any channel noise, \textbf{leading to suboptimal reconstructed results} in the presence of noise, we integrate a noise reduction block (NRB) based on a stack of residual Swin Transformer layers, as mentioned in~\cite{chen2022real} into decoder $G$ to conduct generative image reconstruction.

Building upon the pretrained model and multilevel codebook from Stage 1, as depicted in Fig.~\ref{fig:framework} (Left), \textbf{Stage 2 introduces a physical communication channel} (refer to Section~\ref{sub:Physical_Channe} for details) between transmitter and receiver for physical channel simulation. Then \textbf{the proposed NRB is incorporated after the retrieval noisy feature $\mathbf{\hat{z}}_q$} to refine semantic features, yielding the refined semantic feature $\mathbf{\hat{z}}_r$.

\paragraph{Feature Requantization} To further enhance the quality of $\mathbf{\hat{z}}_r$, we propose requantizing it to obtain $\mathbf{\tilde{z}}_r$ by:
\begin{align}
\mathbf{\tilde{z}}_r = \mathcal{MOC-RVQ}(\mathbf{\hat{z}}_r). 
\end{align}
The intuition behind this lies in the fact that MOC-RVQ is trained in the absence of any noise, establishing itself as a high-quality semantic knowledge base to inject high quality prior for better feature  restoration.
\paragraph{Training Objectives} For NRB finetuning, we first generate the \textbf{noiseless} quantized semantic feature of image $\mathbf{I}$ through $\mathbf{z}_q = \mathcal{MOC-RVQ}(E(\mathbf{I}))$ \textbf{as ground-truth feature.} Then, this ground-truth feature is utilized to calculate the Gram matrix loss and L2 loss for the refined semantic feature $\mathbf{\hat{z}}_r$ by:
\begin{align}
\mathcal{L}_{NR}= \|\psi(\mathbf{\hat{z}}_r)-\psi(sg[\mathbf{z}_q])\|_{2}^{2} + \alpha \|\mathbf{\hat{z}}_r-sg[\mathbf{z}_q]\|_{2}^{2},
\end{align}
where $\psi(\cdot)$ is the flattened version of the Gram matrix of a given feature, and $\alpha$ is set to $0.25$. The first term of $\mathcal{L}_{NR}$, also known as style loss, has been demonstrated to be effective for improved feature recovery~\cite{chen2022real}. Moreover, we incorporate image-level $L_1$ loss for training supervision. It is also noteworthy that \textbf{the autoencoder and codebook trained in Stage 1 remain fixed, and we only finetune the NRB for noise reduction}.

\subsection{Codebook Reordering}%
\label{sub:Reordering}
As discussed in Section~\ref{sec:intro}, a discrepancy exists in the local relationship between code indices and code vectors. In response to this challenge, \textbf{before commencing Stage 2}, we propose a heuristic solution through a codebook reordering (CR) algorithm outlined in Algorithm~\ref{algo:codebook_reordering}. The algorithm systematically searches for the nearest neighbor vector within $\mathcal{C}$, resulting in a new vector sequence, which is then assembled to form a preliminarily sorted codebook. Furthermore, to align with the Gray code mapping process in digital constellation modulation, we also integrate Gray code mapping into our CR algorithm  (optional). The basic idea is to establish proximity relationships under the context of Gray code mapping. For instance, given a 3-bit Gray code sequence represented in decimals as $g=\{0, 1, 3, 2, 6, 7, 5, 4\}$, it becomes apparent that "2" and "6" are neighboring elements. Thus, it is essential to allocate two closely related code vectors to these indices.

\IncMargin{1em}
\begin{algorithm}[t]
    \SetKwData{Left}{left}\SetKwData{This}{this}\SetKwData{Up}{up}
    \SetKwFunction{FindCompress}{FindCompress}\SetKwFunction{Union}{Union}
    \SetKwInOut{Input}{input}\SetKwInOut{Output}{output}

    \Input{A disordered codebook $\mathcal{C}$ of size $N\times n_q$}
    \Output{An ordered codebook $\mathcal{C}^*$ of size $N\times n_q$}
    
    \tcp{Initialization}
    $\mathcal{C}^* \leftarrow \varnothing$ \;
    $\mathbf{s}_0 \leftarrow$ mean of $\mathcal{C}$ along dimension 0 \;
    \For{$i\leftarrow 1$ \KwTo $N$}{
        $\mathbf{s}_i\leftarrow$ the vector closest to $\mathbf{s}_{i-1}$ in $\mathcal{C}$ \;
        Remove $\mathbf{s}_i$ from $\mathcal{C}$ \;
        Append $\mathbf{s}_i$ to $\mathcal{C}^*$ \;
    }
    \tcp{Gray Code Mapping}
    $g \leftarrow$ generate Gray code sequence of size $N$ \;
    $\mathcal{C}^*[g,:] \leftarrow$  $\mathcal{C}^*$ \;
    \caption{Codebook Reordering}
    \label{algo:codebook_reordering}
\end{algorithm}
\DecMargin{1em}

\addtolength{\topmargin}{0.01in}

After codebook reordering, the semantic distance between two adjacent Gray code vectors in the updated codebook $\mathcal{C}^*$ is reduced compared to the previous disordered version. This \textbf{addresses the issue where a small jump in the code index would lead to a significant jump in the semantic vector.}

\section{Experiments}%
\label{sec:Experiments}
In this section, we empirically demonstrate the qualitative and quantitative results of our model, conduct model ablation, and compare our model with different baselines.

\begin{figure*}[t] \centering

\includegraphics[width=0.31\textwidth]{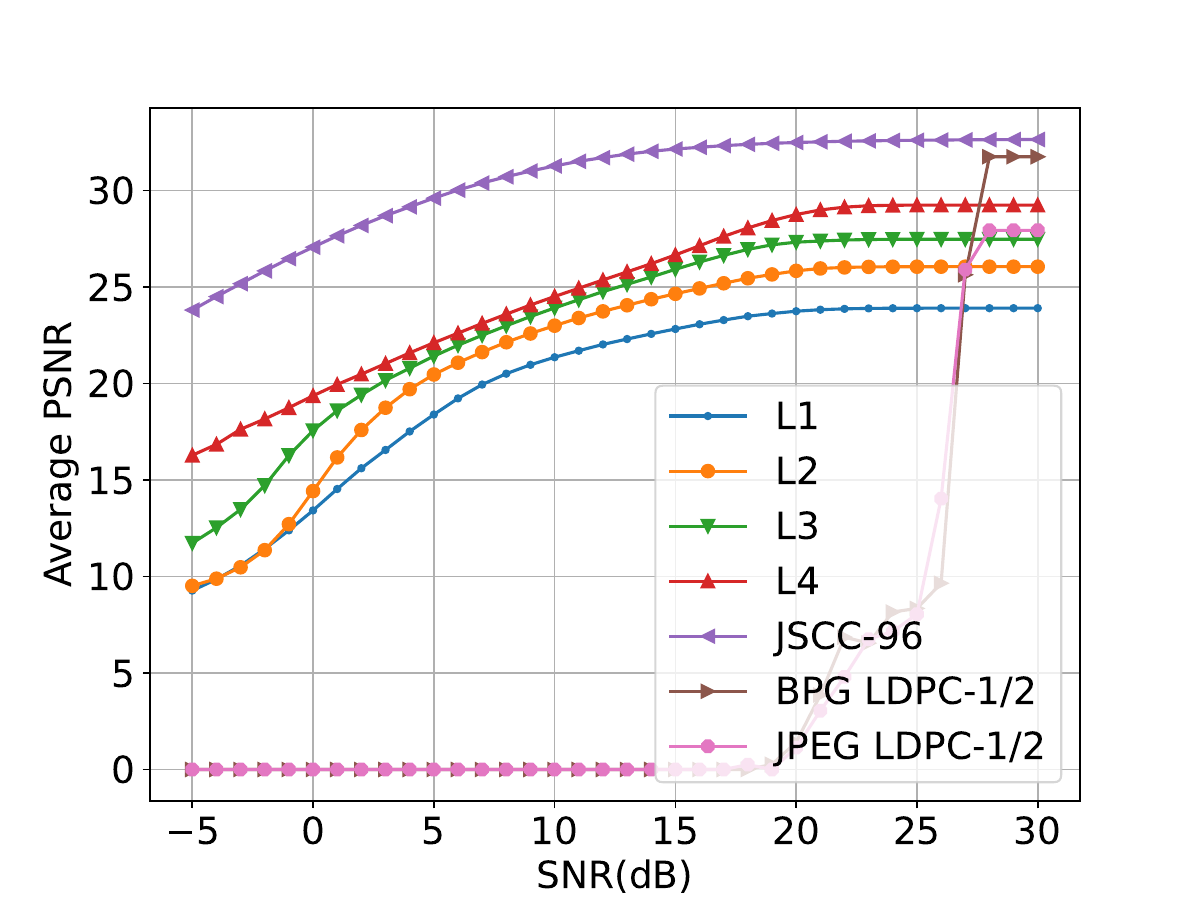}
\includegraphics[width=0.31\textwidth]{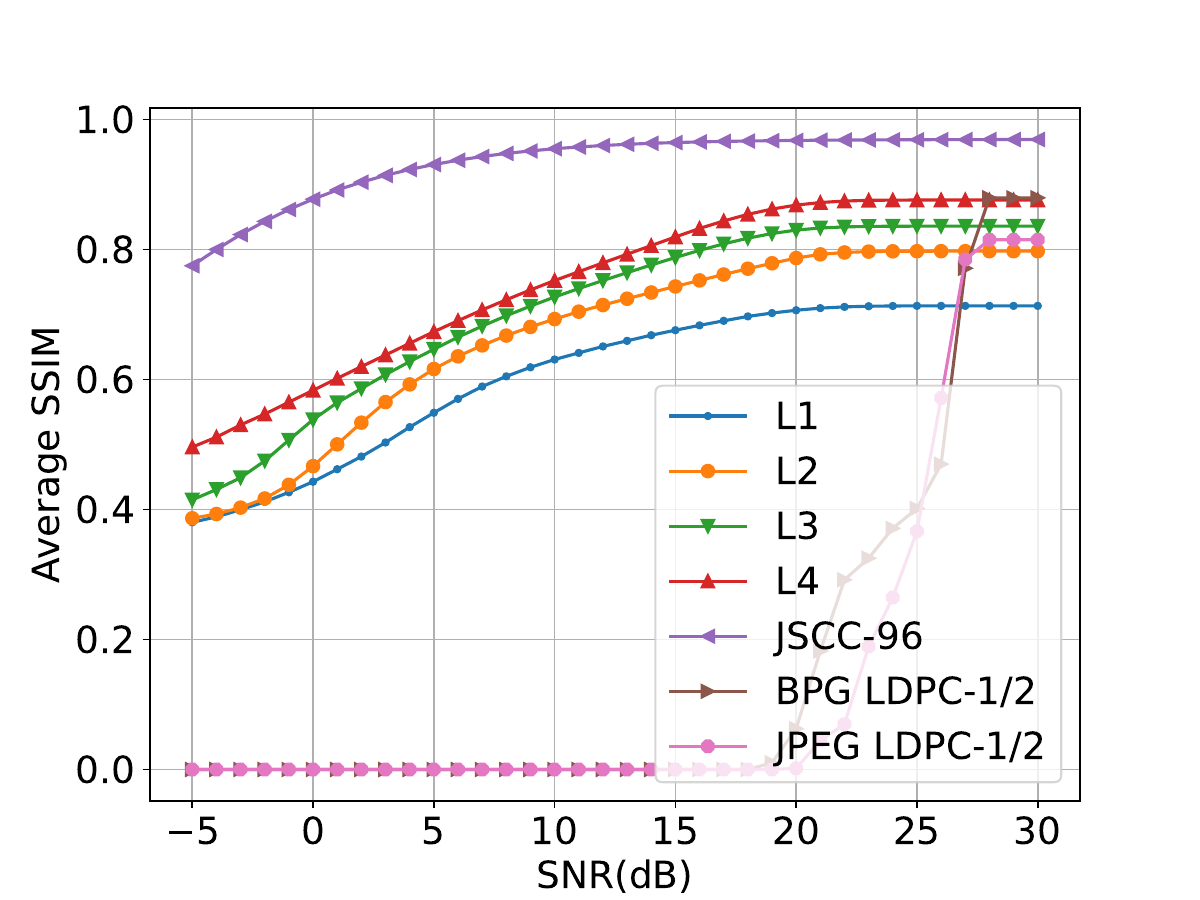}
\includegraphics[width=0.31\textwidth]{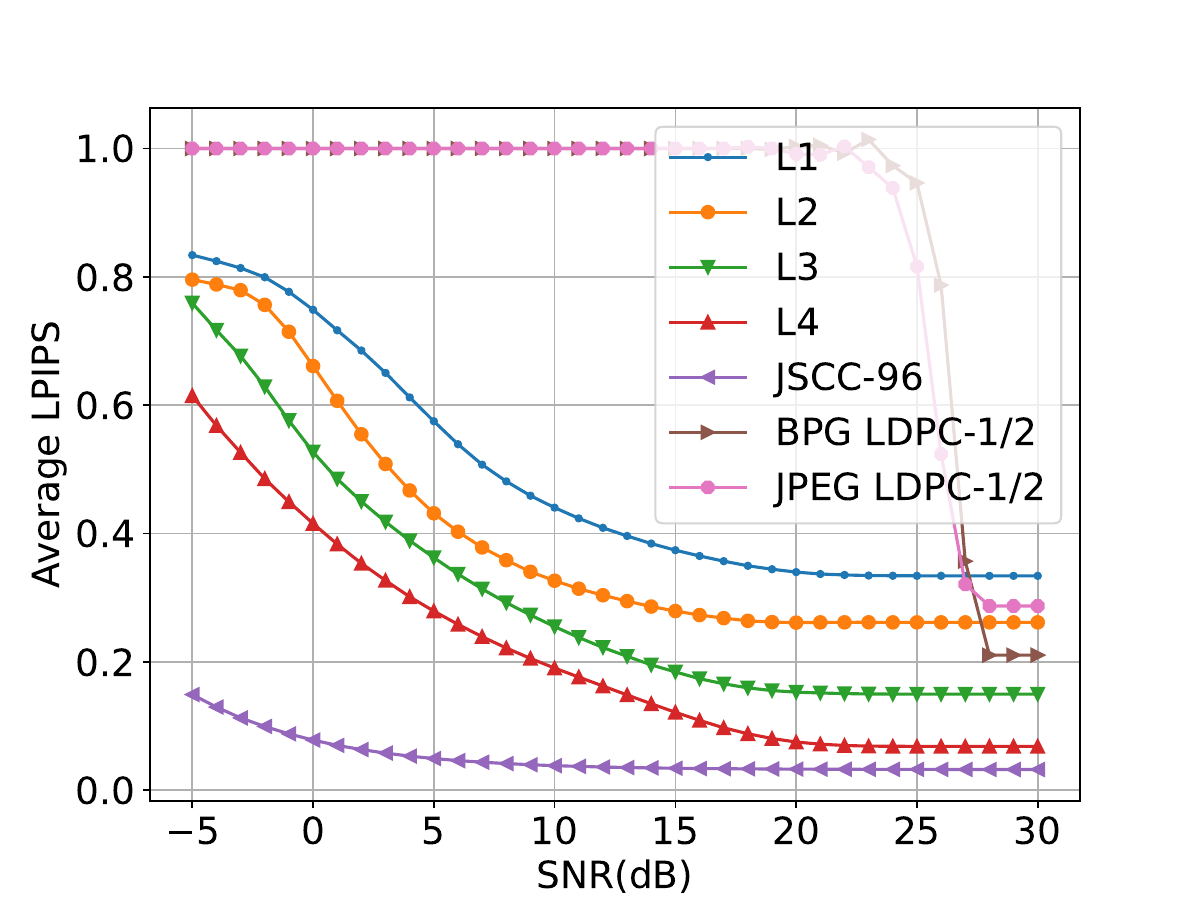}

\makebox[0.31\textwidth]{\scriptsize (a) PSNR}
\makebox[0.31\textwidth]{\scriptsize (b) SSIM}
\makebox[0.31\textwidth]{\scriptsize (c) LPIPS}

\caption{Experimental comparison using PSNR, SSIM, and LPIPS metrics over AWGN channels with SNR from -5 to 30. L1 to L4 correspond to different quantization levels of MOC-RVQ, with L1 as a baseline for the VQGAN-based variant. JSCC-96 is an analog baseline, with a CBR six times higher than L4, and requires analog modulation or full-resolution constellation, which \textbf{hinders compatibility with current digital communication systems.}
} 

\label{fig:main_cmp}
\end{figure*}

\begin{figure*}[t] \centering

\includegraphics[width=0.31\textwidth]{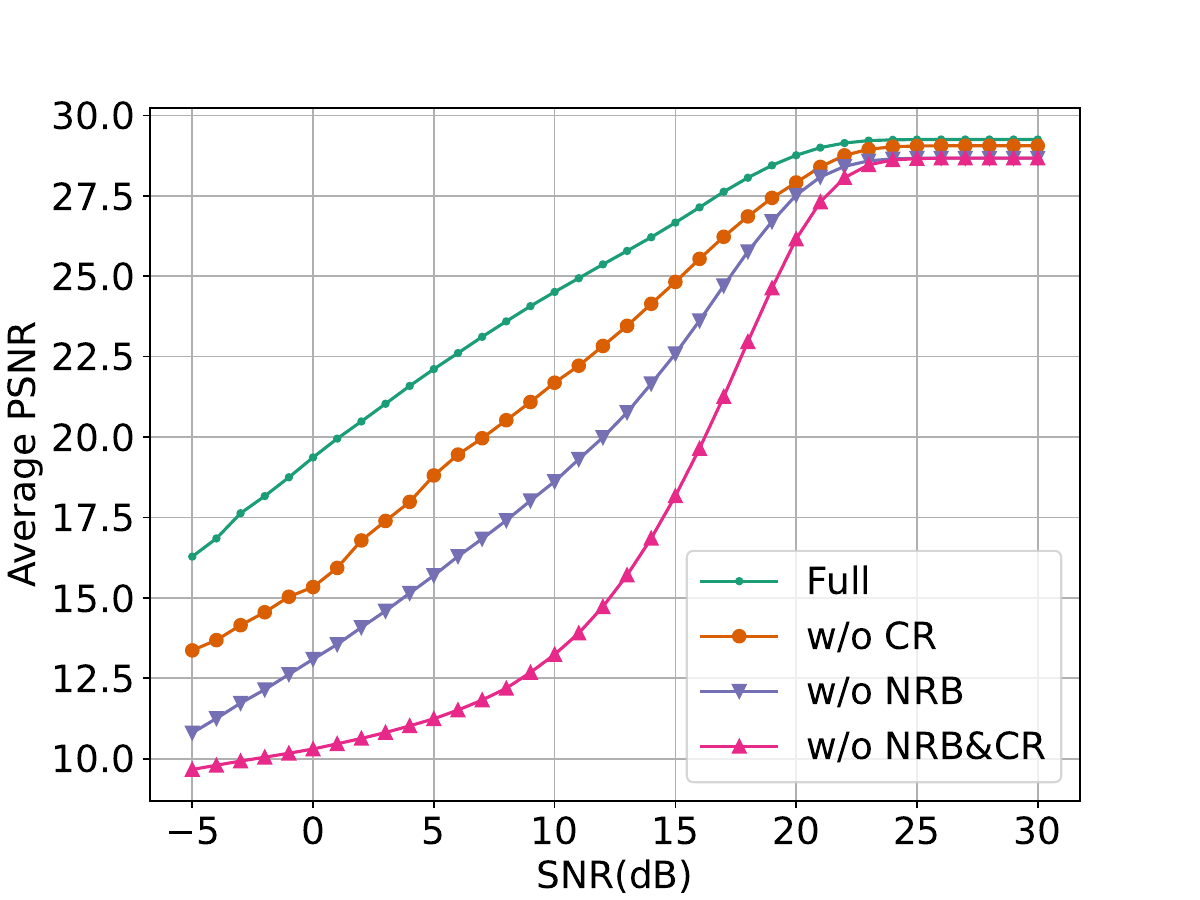}
\includegraphics[width=0.31\textwidth]{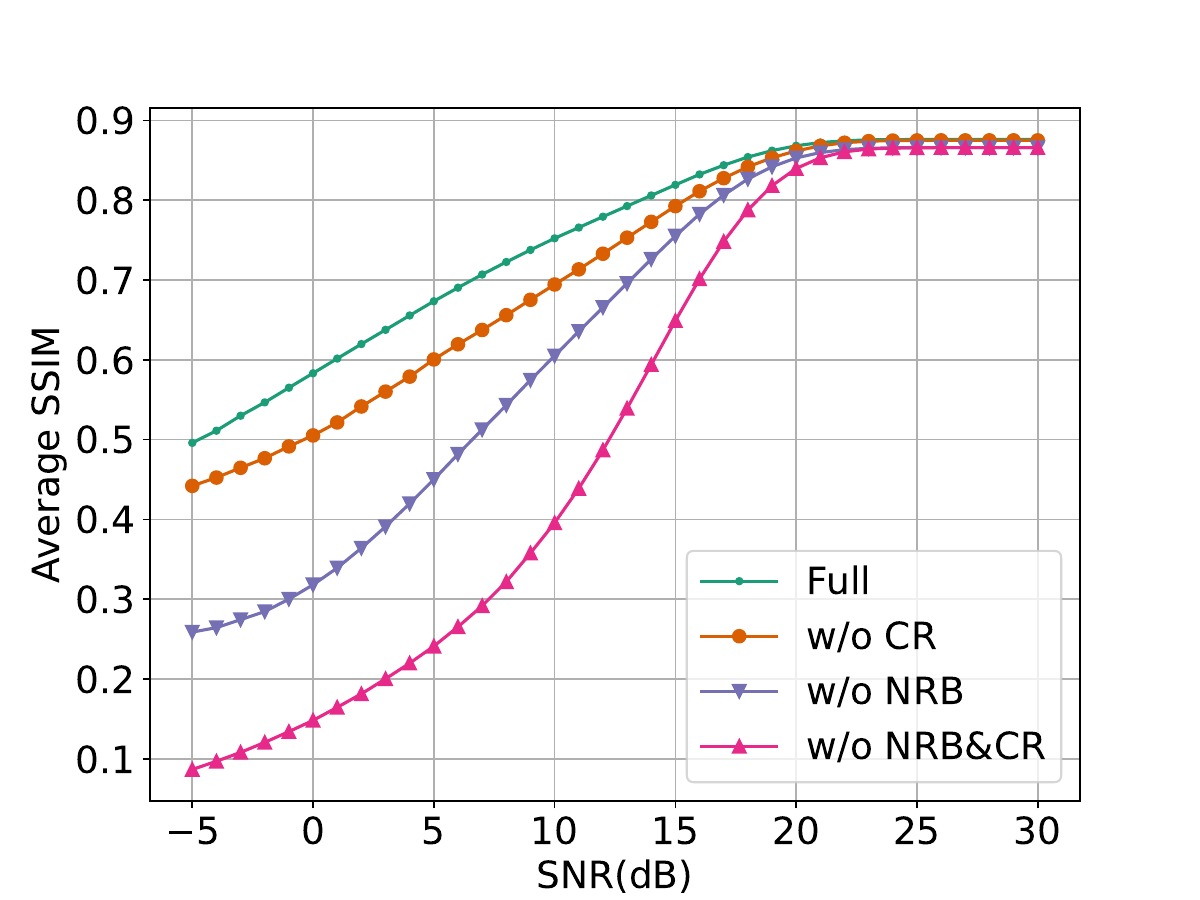}
\includegraphics[width=0.31\textwidth]{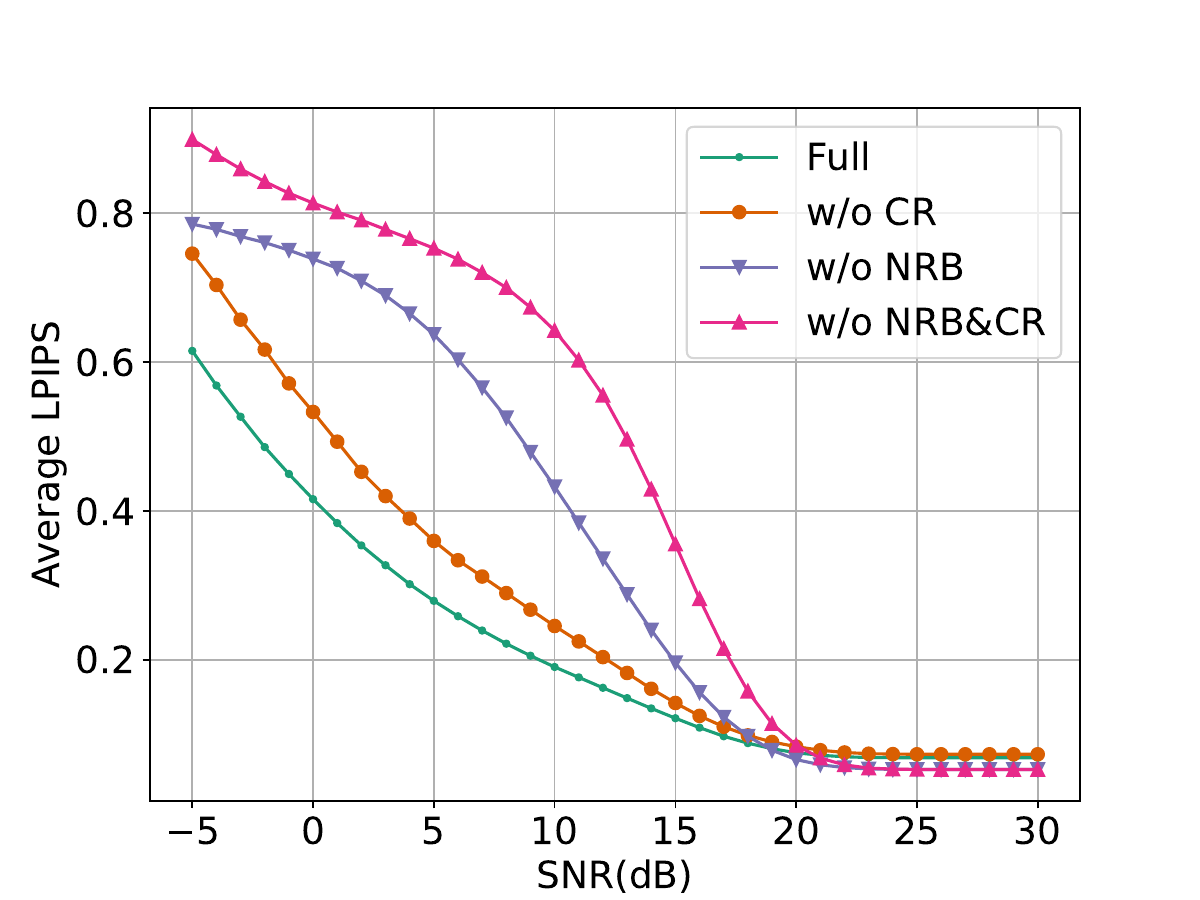}

\makebox[0.31\textwidth]{\scriptsize (a) PSNR}
\makebox[0.31\textwidth]{\scriptsize (b) SSIM}
\makebox[0.31\textwidth]{\scriptsize (c) LPIPS}

\caption{Ablation study to comprehend the influence of our proposed noise reduction block (NRB) and codebook reordering (CR) algorithm. Note that all experiments are conducted under L4 transmission, and 'w/o' denotes 'without'. There are clear gaps between different setting, which verifies the effectiveness of the proposed NRB and CR designs.} 
\label{fig:ab_study}
\end{figure*}

\subsection{Experimental Setups}
\label{sub:Exp_setups}
\paragraph{Model and Dataset} Similar to~\cite{chen2022real}, we utilize the autoencoder from VQGAN \cite{esser2021taming} with a downsampling factor of $f = 8$, meaning that the resolution of the bottleneck feature is $1/f^2$ of the input, as the backbone. For effective model training, we assemble a comprehensive training set by integrating: DIV2K~\cite{div2k}, Flickr2K~\cite{flickr2k}, DIV8K~\cite{gu2019div8k}, and a dataset comprising 10,000 facial images from FFHQ~\cite{ffhq}. Then the training patches are generated by the following steps: (1) Extraction of non-overlapping patches with a resolution of $512 \times 512$; (2) Filtration of patches with minimal texture variation; (3) For the FFHQ dataset, a random resizing operation is applied with scale factors in the range [0.5, 1.0] before cropping. For testing, we evenly sample 25 \textbf{high-resolution images} ($2012 \times 1407$ in average) from the DIV2K validation set to simulate modern image transmission scenarios.

\paragraph{Training Details} We employ an Adam optimizer with $\beta_1=0.9$ and $\beta_2=0.99$, and maintain a learning rate of $0.0001$ throughout the training process. The input image size is randomly cropped to $256 \times 256$ for efficient training with a batch size of 16. The model pretraining stage requires approximately 3 days on 2 GeForce RTX 3090 GPUs, while the NRB finetuning stage also takes about 3 days on the same device.

\paragraph{Channel Settings and Evaluation Metrics} We simulate AWGN channels with SNR varying from -5 to 30 to encompass low, medium, and high-quality channel conditions. Note that the proposed NRB is finetuned under these AWGN channels and all digital models are equipped 64-QAM modulation. For evaluation, PSNR, SSIM, and LPIPS are utilized as metrics for various methods. LPIPS~\cite{lpips} is specifically introduced for evaluating the perceptual quality of generated images under the semantic communication scenarios.

\subsection{Simulations and Discussions}
\label{sub:Simulations}

\begin{figure*}[t] \centering
    \includegraphics[width=0.8\textwidth,height=0.3\textwidth]{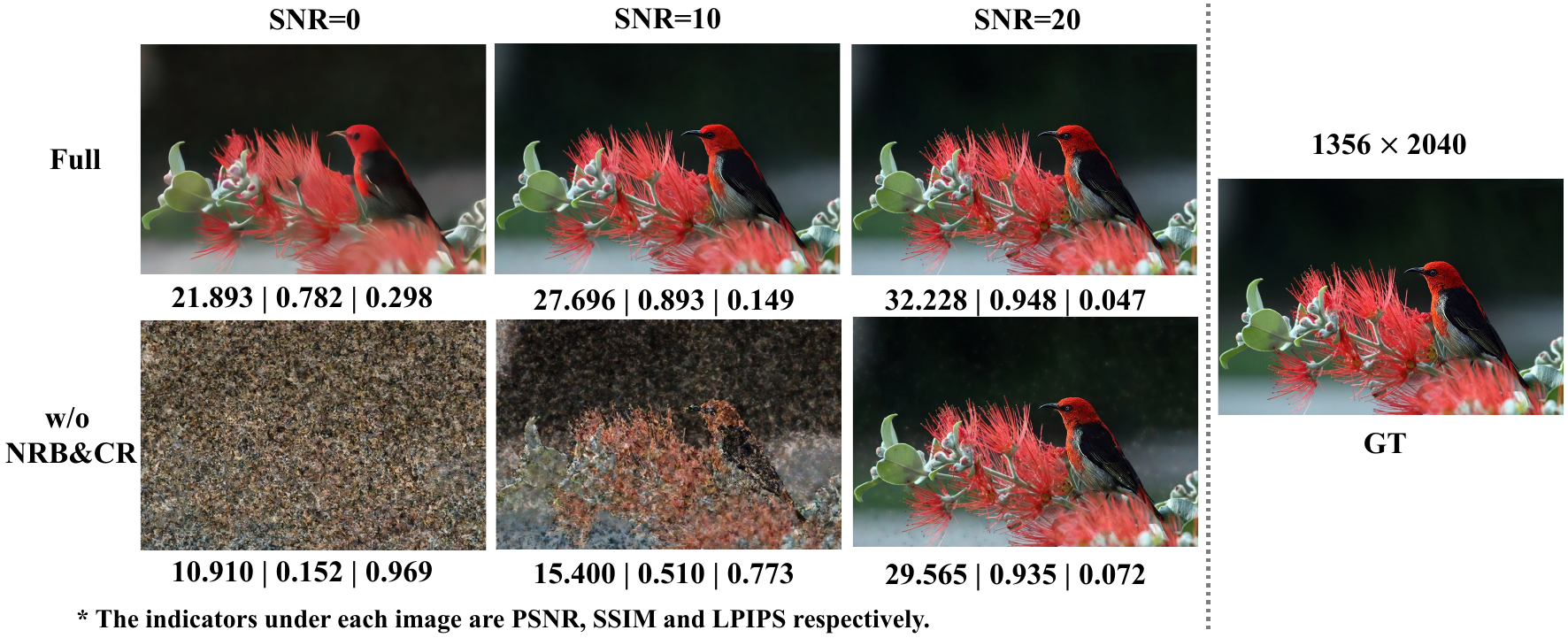}
    \caption{Visualization of removing both the NRB and CR across varying channel conditions. Even in poor channel quality (SNR=0), our full model impressively reconstruct images with clear meaning, demonstrating the robustness of VQ-based semantic communication systems with the assistance of NRB and CR.}
    \label{fig:ab_vis}
\end{figure*}

We apply the better portable graphics (BPG~\cite{bpg_codec}) and the joint picture expert group (JPEG~\cite{wallace1991jpeg}) as digital baselines by setting the number of bits after compression similar to our proposed MOC-RVQ. Notably, both of these image codec formats are susceptible to bit errors, leading to potential image format errors and decoding failures. To mitigate this, we introduce a widely used channel error correction code called Low-Density Parity-Check (LDPC) by setting the LDPC block length and bit rate as 648 bits and 1/2, respectively. Thanks to the self-restoring capability of our proposed method, we do not apply any channel error correction code for MOR-RVQ. 

Additionally, we implement an analog JSCC baseline, referred to as JSCC-96, by retaining only the autoencoder from MOC-RVQ and removing all its digital components. Without loss of generality, we set the number of channels in the bottleneck feature (the encoder output) to 96 for comparison, resulting in a channel bandwidth ratio (CBR) that is six times greater than that of MOC-RVQ-L4. The definition of CBR is as follows:
\begin{equation}
    CBR = \frac{B}{H \times W \times 3},
\end{equation}
where $B$ denotes the number of transmitted complex symbols, and $H,W$ represent the width and height of the input RGB image, respectively. Note that \textbf{the symbols of analog scheme require full-resolution constellation, while the symbols of our method only need few discrete states.} 

From Fig.~\ref{fig:main_cmp}, we can observe that our proposed method exhibits robustness in deteriorating channel quality when compared with traditional image coding, even in the absence of a channel error correction mechanism. Additionally, while the analog scheme JSCC-96 demonstrates commendable performance, it requires a higher CBR and faces compatibility challenges with existing digital communication systems.

Figure~\ref{fig:main_cmp} also shows that reducing the transmission level (from L4 to L1) in our model, which directly decreases the number of transmitted bits, leads to a compromise in reconstruction performance. Notably, the gaps at different levels are clearly visible in the LPIPS metrics. It is also worth mentioning that L1 can serve as a special VQGAN-based digital baseline for comparison.

Additionally, We also conduct ablation study (illustrated in Fig.~\ref{fig:ab_study}) to assess the effects of our proposed NRB and the CR algorithm. We can observe that the proposed NRB plays a crucial role in the reconstruction performance. Simultaneously, the proposed CR algorithm contributes additional gains to the image reconstruction performance.

Figure~\ref{fig:ab_vis} illustrates the visual effect of removing both the NRB and CR. With the assistance of both these designs, even under relatively poor channel quality (SNR=0), our full model can impressively reconstruct images with clear semantic meaning, thus significantly expanding the working conditions of VQ-based semantic communication systems.

\section{Conclusion}
In this work, we propose MOC-RVQ, a novel approach that combines a multi-head octonary codebook (MOC) with residual vector quantization (RVQ) in a two-stage training framework to address the challenges between vector quantization and digital constellation modulation. By leveraging MOC for codebook compression and RVQ for efficient multilevel semantic communication, our method surpasses traditional techniques like BPG and JPEG, and achieves performance comparable to analog JSCC. Furthermore, the two-stage framework, enhanced by the NRB design and CR algorithm, significantly improves the system’s ability to recover high-quality semantic features under channel noise. While effective, MOC-RVQ is specifically optimized for 64-QAM modulation, suggesting the need for codebook adjustments and retraining for other modulation schemes, providing avenues for future research.

\bibliographystyle{IEEEtran}
\bibliography{ref}


\end{document}